\documentclass[10pt,twocolumn,letterpaper]{article}

\usepackage{cvpr}              

\usepackage[dvipsnames, table]{xcolor}
\usepackage{xcolor}
\usepackage{ifthen} 
\usepackage{lipsum}

\usepackage{array}
\usepackage{colortbl}
\usepackage{dsfont}
\usepackage{amsmath}
\usepackage{multirow}
\usepackage{graphicx}
\usepackage{tikz}
\usepackage{float}
\usepackage{caption}
\usepackage{svg}
\usepackage{adjustbox}

\definecolor{cvprblue}{rgb}{0.21,0.49,0.74}
\usepackage[pagebackref,breaklinks,colorlinks,citecolor=cvprblue]{hyperref}

\def\paperID{208} 
\def\confName{3DV\xspace}
\def\confYear{2026\xspace}

\title{PI3DETR: Parametric Instance Detection of 3D Point Cloud Edges With a Geometry-Aware 3DETR}

\author{
Fabio F. Oberweger\thanks{Equal contribution.} \quad
Michael Schwingshackl\footnotemark[1] \quad
Vanessa Staderini \\
AIT Austrian Institute of Technology\\Center for Vision, Automation \& Control\\
{\tt\small \{fabio.oberweger, michael.schwingshackl, vanessa.staderini\}@ait.ac.at}
}
\begin{document}
\twocolumn[{
  \maketitle
  \begin{center}
    \includegraphics[width=\textwidth]{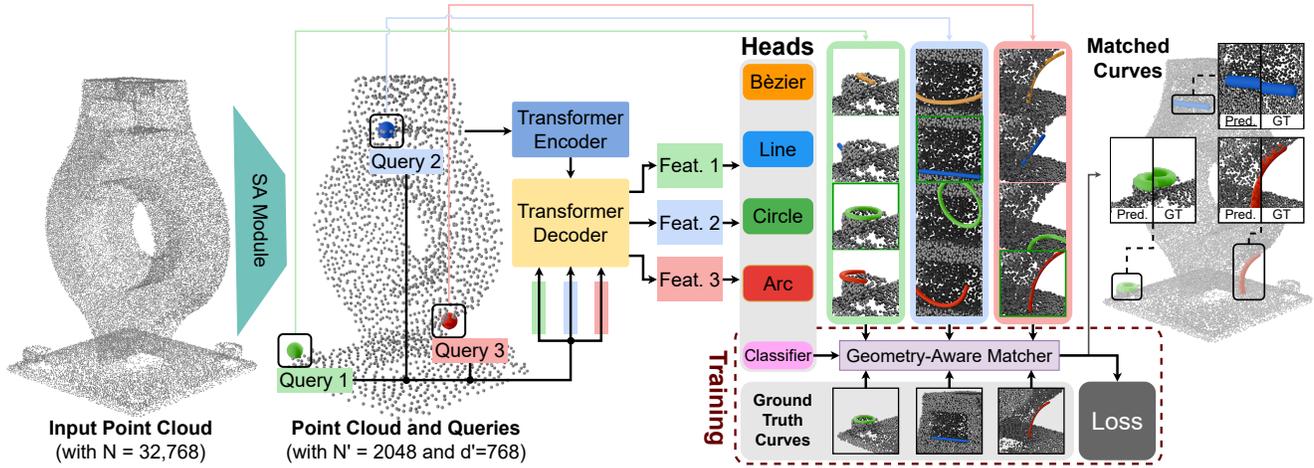}
    {\captionsetup{hypcap=false}
     \captionof{figure}{
     PI3DETR is an end-to-end pipeline that takes a point cloud as input. Like 3DETR \cite{misra2021end}, it uses an SAModule \cite{qi2017pointnet++} to sample points and generate queries, and a Transformer \cite{vaswani2017attention} to extract query-specific features. Multiple heads predict parametric curves (cubic Bézier, line segments, circle, arc) with associated parameters, while a geometry-aware matcher aligns predictions with ground truth, removing the need for intermediate representations or post-processing. This visualization shows three sample queries for clarity.
     }
     \label{fig:method_overview}}
  \end{center}
}]
\footnotetext[1]{Equal contribution.}
\begin{abstract}
We present PI3DETR, an end-to-end framework that directly predicts 3D parametric curve instances from raw point clouds, avoiding the intermediate representations and multi-stage processing common in prior work. Extending 3DETR, our model introduces a geometry-aware matching strategy and specialized loss functions that enable unified detection of differently parameterized curve types, including cubic Bézier curves, line segments, circles, and arcs, in a single forward pass. Optional post-processing steps further refine predictions without adding complexity. This streamlined design improves robustness to noise and varying sampling densities, addressing critical challenges in real world LiDAR and 3D sensing scenarios. PI3DETR sets a new state-of-the-art on the ABC dataset and generalizes effectively to real sensor data, offering a simple yet powerful solution for 3D edge and curve estimation. Code: \href{https://github.com/fafraob/pi3detr}{https://github.com/fafraob/pi3detr}.
\end{abstract}
\section{Introduction}
\label{sec:1_introduction}

\begin{figure}
    \centering
    \includegraphics[width=\linewidth]{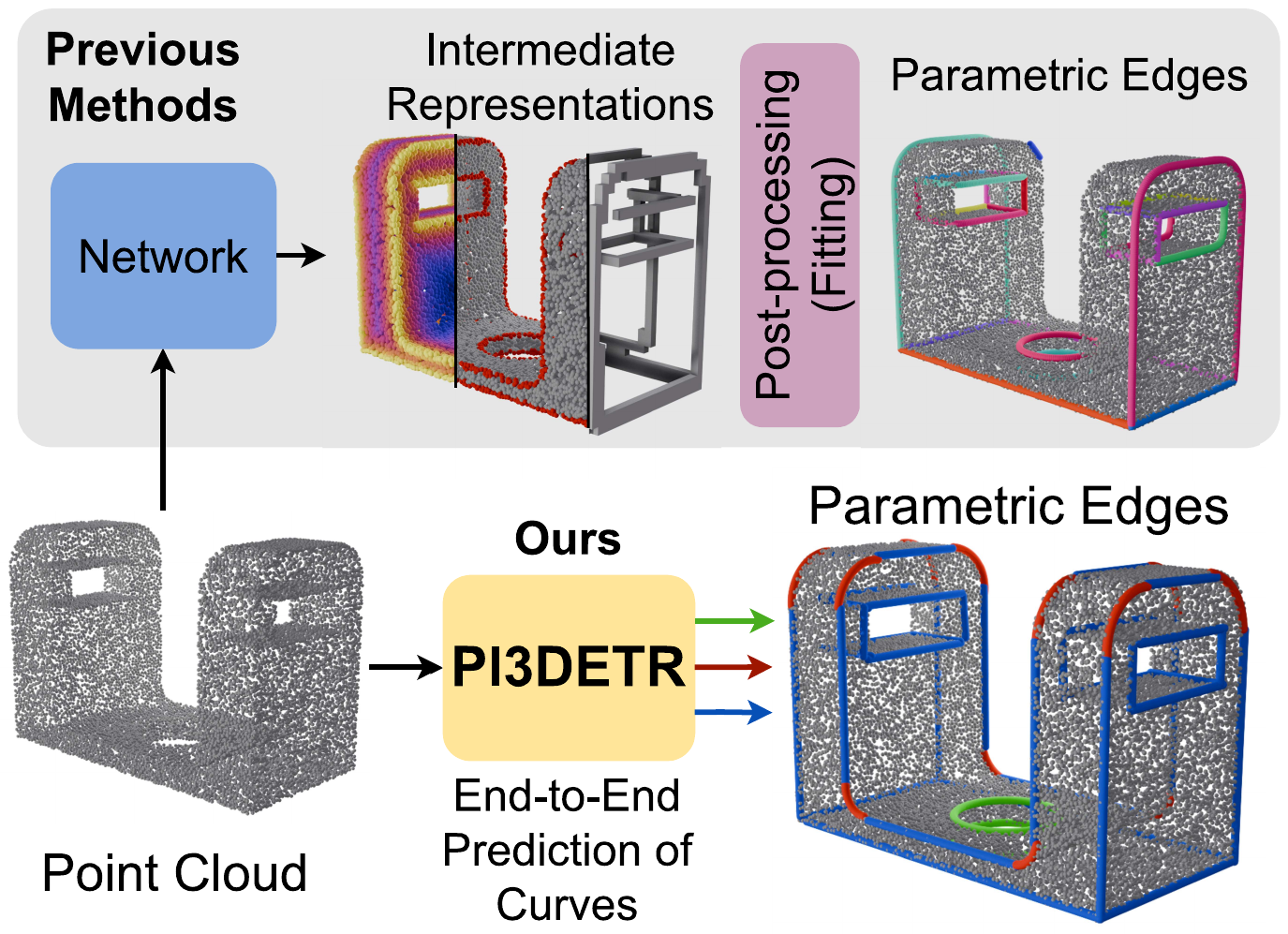}
    \caption{
    Unlike methods that rely on intermediate representations (distance fields, segmentations, voxels) and post-fitting, PI3DETR directly predicts parametric edge curves in an end-to-end manner.}
    \label{fig:method_comparison}
\end{figure}

Estimating 3D edges in point clouds is not only a compelling theoretical challenge but also a task with significant practical impact. In robotic vision, for example, edge information can be used to optimize grasp planning by determining precisely how and where an object should be gripped. In manufacturing, wireframe extraction enables more effective quality control, while in other domains, such as autonomous driving, CAD reconstruction, and various 3D data processing pipelines, edges serve as essential structural cues. In many of these applications, LiDAR (light detection and ranging) and other 3D sensing technologies are employed to capture complex objects and environments. However, the resulting point clouds are rarely flawless as they often contain noise, exhibit non-uniform sampling densities, and suffer from occlusion-induced gaps. These imperfections make the development of robust and adaptable edge detection methods critical for ensuring reliable performance across diverse real-world scenarios.

Hence, we introduce PI3DETR, a streamlined end-to-end architecture that directly predicts parametric curves from raw point clouds, avoiding the complexity and modularity inherent in prior approaches. Existing methods often rely on intermediate representations such as voxel grids \cite{zhu2023nerve}, range images \cite{matveev2022def}, or 2D sketches \cite{wang2025point2primitive}, which introduce additional processing steps and impose assumptions on the input space. Others incorporate post-processing stages to fit curves to precomputed segmentations \cite{wang2020pie} or piece-wise linear primitives \cite{zhu2023nerve}, or construct curve graphs using multi branch decoders \cite{guo2022complexgen}. In contrast, PI3DETR employs a unified architecture that directly infers parametric curve instances from point clouds, eliminating auxiliary representations, handcrafted pipelines, and multi stage reasoning. As illustrated in \cref{fig:method_comparison}, our approach simplifies the overall pipeline while expanding capability.
To achieve this, we adapt the 3DETR framework \cite{matveev2022def} and extend it with a novel geometry-aware matching strategy that enables joint detection of multiple curve types including cubic Bézier curves, line segments, circles, and arcs in a single forward pass. We further introduce tailored loss functions specifically designed for 3D edge estimation and propose optional post-processing steps that yield additional performance gains without being integral to the pipeline.

Another important aspect of our work is robustness to variations in point cloud sampling and noise. Both, our experiments and prior studies \cite{zhu2023nerve,wang2020pie}, show that performance can degrade substantially when sampling densities change or noise is added. We hypothesize that our simplified end-to-end architecture enhances robustness by eliminating the dependency on intermediate representations or highly parameterized post-processing methods.

We evaluate PI3DETR on the ABC dataset \cite{koch2019abc} and compare it to the state-of-the-art approach NerVE \cite{zhu2023nerve}. Our results show that PI3DETR surpasses NerVE both qualitatively and quantitatively on this challenging benchmark. Furthermore, we collect a small dataset of real world sensor data to demonstrate the practical applicability of our method. Our main contributions are:
\begin{itemize}
\item PI3DETR, a streamlined architecture that directly predicts 3D curve instances with different parameterizations from raw point clouds.
\item A geometry-aware prediction and matching strategy that enables detection of cubic Bézier curves, line segments, circles, and arcs within a single model.
\item A simple yet effective optional post-processing procedure for parametric curve prediction.
\item State-of-the-art performance on the ABC dataset for 3D parametric curve detection.
\end{itemize}
\section{Related Work}
\label{sec:2_related_work}

As noted by Zhu \etal \cite{zhu2023nerve}, a common strategy for parametric curve reconstruction is to first detect geometric primitives such as corners and edges and then then fit curves in a separate post-processing stage \cite{wang2020pie,guo2022complexgen,matveev2022def,liu2021pcwf}. PIE-NET \cite{wang2020pie} follows this paradigm using a PointNet++ \cite{qi2017pointnet++} backbone to classify points into corner, edge, or background classes, after which a secondary network predicts curve proposals between detected corner pairs and estimates their parameters. The method supports lines, circles, and B-splines, and is trained on the ABC dataset \cite{koch2019abc}.

EC-Net \cite{yu2018ec} addresses a related problem of edge-aware point cloud upsampling by jointly learning point consolidation and edge preservation. It predicts an edge confidence score to guide new point placement along sharp features, producing denser point clouds with crisper edges and corners compared to prior upsampling methods.

ComplexGen \cite{guo2022complexgen} discretizes point clouds into sparse voxels and applies a 3D CNN \cite{choy20194d} to extract features, which are decoded via three DETR \cite{carion2020end} decoder modules for vertices, edges, and faces. The predictions are then assembled into a CAD model by solving an integer linear program.

DEF \cite{matveev2022def} processes dense point clouds via a CNN U-Net \cite{ronneberger2015u} on range images to predict per point distance-to-feature fields. After thresholding, it estimates corners, segments curves, and constructs a connectivity graph that is optimized to produce parametric lines, circles, and B-splines.

Unlike keypoint based approaches, NerVE~\cite{zhu2023nerve} performs volumetric edge detection by combining PointNet++ with a 3D CNN \cite{choy20194d} to produce a voxel feature grid from the input point cloud. Three encoder branches predict voxel occupancy by edge points, their 3D coordinates, and their connectivity to neighbors. From these predictions, piece-wise linear curves (PWLs) are assembled and refined through a dedicated post-processing step to form structured boundary representations.

Since the introduction of DETR~\cite{carion2020end} for 2D object detection, its transformer-based architecture with set prediction has been adapted to 3D tasks. 3DETR~\cite{misra2021end} extends DETR to 3D bounding box detection directly on point clouds, using a set aggregation module~\cite{qi2017pointnet++} for downsampling followed by Transformer \cite{vaswani2017attention} encoder and decoder blocks. In contrast to DETR’s learned queries, 3DETR uses actual point locations as queries, grounding predictions in the input geometry.

Point2Primitive~\cite{wang2025point2primitive} addresses curve extraction as part of CAD reconstruction. It applies PointNet++ to cluster extrusion regions, then uses an adapted DETR to detect 2D sketch primitives such as lines, circles, and arcs within each extrusion. These sketches are combined with extrusion operations to reconstruct complete CAD models.
\section{Method}
\label{sec:3_method}

PI3DETR takes a 3D point cloud as input and predicts 3D edges in the form of different curve types in a parameterized way. In the following, we describe the training data, the parameterization, and the model architecture.

\subsection{Dataset and Ground Truth}
\label{sec:dataset}
The ABC dataset \cite{koch2019abc} is a standard benchmark for parametric edge detection \cite{wang2020pie,matveev2022def,zhu2023nerve} and CAD reconstruction \cite{guo2022complexgen,wang2025point2primitive}, providing paired CAD and mesh representations. Following prior work \cite{wang2020pie,matveev2022def,zhu2023nerve}, we focus on sharp edges in the CAD models and apply the filtering protocol in \cref{apx:dataset} to remove erroneous or inconsistent entries. We further retain only models whose curves can be represented as cubic Bézier curves, line segments, circles, or arcs, and contain at most $100$ curves. This yields $10{,}240$ models, split into $80\%$ training, $10\%$ validation, and $10\%$ test. To create our 3D point cloud dataset $\mathcal{P}$, we sample $32{,}768$ surface points for each mesh in the remaining dataset. Triangle point picking~\cite{trimesh} is used as the sampling method to promote generalization, in contrast to approaches that primarily rely on mesh vertices~\cite{zhu2023nerve}.

\paragraph{Parametrization and ground truth definition.}
Let $P \in \mathcal{P}$ denote an arbitrary point cloud from the dataset, and let $M$ be the number of parametric curves associated with $P$. To adapt the curve representations of the CAD model underlying $P$ for the learning task, we extract curve vertices and fit parametric primitives to each segment.

\begin{figure}
    \centering
    \includegraphics[width=1\linewidth]{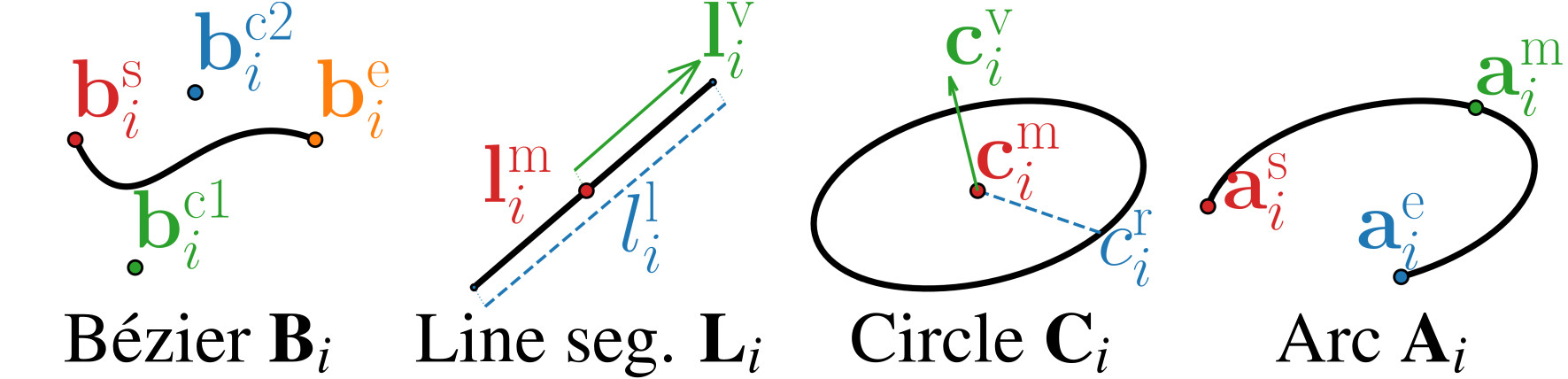}
\caption{Parametric representations of the four curve types used for prediction: Bézier ($B_i$), Line Segment ($L_i$), Circle ($C_i$), and Arc ($A_i$).}
    \label{fig:curve_representations}
\end{figure}

We denote each ground truth curve in $P$ by an index $i \in \{1, \dots, M\}$. Each curve is assigned a class label $c_i \in \{1, 2, 3, 4\}$ indicating its type: $1$ for a cubic Bézier curve, $2$ for a line segment, $3$ for a circle, and $4$ for an arc. The parameters for each curve type are defined as follows, and are illustrated in \cref{fig:curve_representations}:
A cubic Bézier curve $\mathbf{B}_i=[\mathbf{b}^\text{s}_i, \mathbf{b}^\text{c1}_i, \mathbf{b}^\text{c2}_i, \mathbf{b}^\text{e}_i]$ ($c_i=1$) is defined by four ordered control points $\mathbf{b}^\text{s}_i, \mathbf{b}^\text{c1}_i, \mathbf{b}^\text{c2}_i, \mathbf{b}^\text{e}_i \in \mathbb{R}^3$, corresponding to the start point, two control points, and the end point, respectively.
A line segment $\mathbf{L}_i = [\mathbf{l}^{\text{m}}_i, \mathbf{l}^{\text{v}}_i, l^{\text{r}}_i]$ ($c_i=2$) is parameterized by a mid-point $\mathbf{l}^{\text{m}}_i \in \mathbb{R}^3$, a unit direction vector $\mathbf{l}^{\text{v}}_i \in \mathbb{R}^3$, and a length $l^{\text{r}}_i \in \mathbb{R}^+$.
A circle $\mathbf{C}_i = [\mathbf{c}^{\text{m}}_i, \mathbf{c}^{\text{v}}_i, c^{\text{r}}_i]$ ($c_i=3$) is defined by its mid-point $\mathbf{c}^{\text{m}}_i \in \mathbb{R}^3$, a unit normal vector $\mathbf{c}^{\text{v}}_i \in \mathbb{R}^3$ with $\|\mathbf{c}^{\text{v}}_i\| = 1$, and a radius $c^{\text{r}}_i \in \mathbb{R}^+$. 
Finally, an arc $\mathbf{A}_i = [\mathbf{a}^{\text{s}}_i, \mathbf{a}^{\text{m}}_i, \mathbf{a}^{\text{e}}_i]$ ($c_i=4$) is specified by three ordered points $\mathbf{a}^{\text{s}}_i, \mathbf{a}^{\text{m}}_i, \mathbf{a}^{\text{e}}_i \in \mathbb{R}^3$, denoting the arc's start point, mid-point, and end point.

All curve parameters and input point clouds are uniformly normalized with respect to the longest axis of the mesh bounding box, preserving the geometric proportions of the shapes (e.g., preventing circles from being distorted into ellipses). After normalization, the point clouds, the point-, and vector-based curve parameters lie within $[-1,1]^3$.

\subsection{PI3DETR Overview}
PI3DETR builds upon 3DETR \cite{misra2021end} and extends its functionality to predict geometric objects of different shapes, in our case parametric curves, by incorporating a geometry-aware matching framework. The main differences from 3DETR lie in the design of the prediction heads, the loss functions, and the matching strategy, each of which is discussed in the corresponding paragraphs.

\paragraph{3DETR core.}

The input to PI3DETR is a point cloud $P \subset \mathbb{R}^3$ of $N$ points with 3D coordinates only (no normals or colors). First, we use a set-aggregation module (SAModule) \cite{qi2017pointnet++} with farthest point sampling (FPS) \cite{qi2017pointnet++} to downsample to $N'$ points, ensuring even spatial coverage. For each downsampled point, the SAModule extracts a $d$-dimensional local feature, which is linearly projected to $d'$ and processed by Transformer encoder–decoder blocks \cite{vaswani2017attention}. Unlike 3DETR, we apply layer normalization \emph{after} each block’s operations instead of using pre-norm~\cite{klein2017opennmt}.
The encoder produces $N'$ point features, which the decoder processes together with $K$ query embeddings $\{q^\mathrm{f}_j\}_{j=1}^K$ to produce $K$ output embeddings $\{o^\mathrm{f}_j\}_{j=1}^K$ for curve prediction. Following Misra~\etal \cite{misra2021end}, we use non-parametric queries obtained by FPS from the input cloud. Each query point $q_j$ is encoded via sine positional encoding \cite{vaswani2017attention} and projected with a feed-forward network (FFN) to form $q^\mathrm{f}_j$. The same positional encoding is applied to encoder outputs before the decoder, preserving spatial information lost during feature extraction.

\paragraph{Prediction feed-forward networks (FFNs).}
Following the parameterization in \cref{sec:dataset}, we distinguish between two prediction types: 3D point regression (control points, centers, endpoints) and scalar regression (length, radius). Each prediction head is implemented as a FFN parameterized by $\boldsymbol{\theta}^{(\cdot)}$ and takes a decoder output embedding $o^\text{f}_j \in \{o^\text{f}_j\}_{j=1}^K$ as input.

For Bézier curves, $\boldsymbol{\theta}^{\mathbf{B}}$ maps $o^\text{f}_j$ to four control points $\hat{\mathbf{B}}_j = [\hat{\mathbf{b}}^\text{s}_j, \hat{\mathbf{b}}^\text{c1}_j, \hat{\mathbf{b}}^\text{c2}_j, \hat{\mathbf{b}}^\text{e}_j] \in \mathbb{R}^{4\times 3}$. 
For line segments, $\boldsymbol{\theta}^{\mathbf{L}}$ outputs the midpoint $\hat{\mathbf{l}}^\text{m}_j \in \mathbb{R}^3$ and unit direction $\hat{\mathbf{l}}^\text{v}_j \in \mathbb{R}^3$, while $\boldsymbol{\theta}^{\ell}$ predicts the length $\hat{l}^\text{r}_j \in \mathbb{R}$. 
For circles, $\boldsymbol{\theta}^{\mathbf{C}}$ returns the center $\hat{\mathbf{c}}^\text{m}_j \in \mathbb{R}^3$ and unit normal $\hat{\mathbf{c}}^\text{v}_j \in \mathbb{R}^3$, and $\boldsymbol{\theta}^{r}$ estimates the radius $\hat{c}^\text{r}_j \in \mathbb{R}$.
We summarize the line segment and circle predictions as $\hat{\mathbf{L}}_j=[\hat{\mathbf{l}}^\text{m}_j, \hat{\mathbf{l}}^\text{v}_j, \hat{l}^\text{r}_j]$ and $\hat{\mathbf{C}}_j=[\hat{\mathbf{c}}^\text{m}_j, \hat{\mathbf{c}}^\text{v}_j, \hat{c}^\text{r}_j]$, respectively.
For arcs, $\boldsymbol{\theta}^{\mathbf{A}}$ predicts the start point, mid-point, and end point $\hat{\mathbf{A}}_j = [\hat{\mathbf{a}}^\text{s}_j, \hat{\mathbf{a}}^\text{m}_j, \hat{\mathbf{a}}^\text{e}_j] \in \mathbb{R}^{3\times 3}$.
A classification head $\boldsymbol{\theta}^{\text{cls}}$ maps $o^\text{f}_j$ to $\hat{p}_j \in [0,1]^5$ over the set \{no-object, cubic Bézier, line segment, circle, arc\}, selecting the curve type at inference. PI3DETR computes parameters for all curve types from every $o^\text{f}_j$, while a geometry-aware matcher assigns gradients only to the heads corresponding to the ground-truth type during training.

For simplicity, we describe all positions as if they were predicted directly. In practice, each point is obtained as an offset from its associated query location $q_j$, which stabilizes learning and improves localization accuracy.

\paragraph{Geometry-aware matching.}
We match predicted 3D curves to the ground truth using a cost-minimal bipartite assignment. Each decoder output $o_j^\mathrm{f} \in \{o^\text{f}_j\}_{j=1}^K$ predicts parameters for all four curve types, producing $4 \cdot K$ curve hypotheses. For a prediction $j \in \{1,\dots,K\}$ and ground truth $i \in \{1,\dots,M\}$, the matching cost is
\begin{align}
    \mathcal{C}_\text{match}(j, i) = -\hat{p}_j(c_i) + \mathcal{L}_\text{param}(j, i),
\end{align}
where $-\hat{p}_j(c_i)$ is the negative class probability of query $j$ being class $c_i$ \cite{carion2020end} and $\mathcal{L}_\text{param}$ is a geometry-specific parameter loss:
\begin{align}
\label{eq:loss_param}
\begin{split}
    \mathcal{L}&_\text{param}(j, i) = \\
    &\mathds{1}_{\{c_i=1\}} 
    \mathcal{L}_{\text{seq}}(
    \hat{\textbf{B}}_j, 
    \textbf{B}_i) +
    \mathds{1}_{\{c_i=2\}} 
    \mathcal{L}_\text{hybrid}
    (\hat{\mathbf{L}}_j,
    \textbf{L}_i
    ) + \\
    & \mathds{1}_{\{c_i=3\}} 
    \mathcal{L}_\text{hybrid}
    (\hat{\mathbf{C}}_j, 
    \mathbf{C}_i,
    ) +
    \mathds{1}_{\{c_i=4\}} 
    \mathcal{L}_{\text{seq}}
    (\hat{\textbf{A}}_j, \textbf{A}_i).
\end{split}
\end{align}
For cubic Bézier curves and arcs parameterized as ordered point sequences, we adopt an order-invariant $\ell_1$-loss \cite{polyroad} to account for direction ambiguity:
\begin{align}
\label{eq:loss_sequence}
\mathcal{L}_{\text{seq}}(\hat{\mathbf{X}}, \mathbf{X}) = \min \{ \| \hat{\mathbf{X}} - \mathbf{X} \|_1, \, \| \mathrm{rev}(\hat{\mathbf{X}}) - \mathbf{X} \|_1 \},
\end{align}
Here, $\mathrm{rev}(\cdot)$ reverses the point order. For example, $\mathrm{rev}(\hat{\mathbf{B}})=[\hat{\mathbf{b}}^\mathrm{e},\hat{\mathbf{b}}^\mathrm{c2},\hat{\mathbf{b}}^\mathrm{c1},\hat{\mathbf{b}}^\mathrm{s}]$.

Lines $\mathbf{L}=[\mathbf{l}^m,\mathbf{l}^v,l^r]$ and circles $\mathbf{C}=[\mathbf{c}^m,\mathbf{c}^v,c^r]$ share a hybrid point–vector–scalar representation (midpoint/center, direction/normal, length/radius). The vector is sign-ambiguous, so we evaluate both orientations for this loss:
\begin{align}
\label{eq:loss_line_circle}
\begin{split}
    \mathcal{L}_\text{hybrid}(\hat{\mathbf{X}}, \mathbf{X}) = \min \{ 
    \| (\hat{\mathbf{x}}^\text{m}, \hat{\mathbf{x}}^\text{v}) - (\mathbf{x}^\text{m}, \mathbf{x}^\text{v}) \|_1, \, \\
    \| (\hat{\mathbf{x}}^\text{m}, \hat{\mathbf{x}}^\text{v}) - (\mathbf{x}^\text{m}, -\mathbf{x}^\text{v}) \|_1 
    \} + \| \hat{x}^\text{r} - x^\text{r} \|_1 \,.
\end{split}
\end{align}
Following \cite{carion2020end,misra2021end}, we solve for the minimal-cost permutation $\sigma \in \mathfrak{S}_K$ via the Hungarian algorithm \cite{kuhn1955hungarian}, assigning each $i \in \{1,\dots,M\}$ to a unique $j=\sigma(i)$. The remaining $K-M$ predictions are matched to no-object entries ($c_{\sigma(i)}=0$). More details regarding the matching are provided in \ref{apx:matching}.

\begin{figure}[b]
    \centering
    \begin{subfigure}{1\linewidth}
        \centering
        \includegraphics[width=\linewidth]{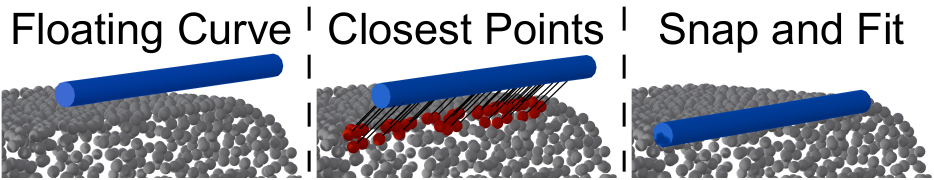}
        \caption{Snap \& Fit (S\&F): realigning slightly offset curves by finding the closest points in the point cloud and refitting the curve.}
        \label{fig:pp_snap}
    \end{subfigure}
    
    \vspace{0.5em} 
    
    \begin{subfigure}{1\linewidth}
        \centering
        \includegraphics[width=\linewidth]{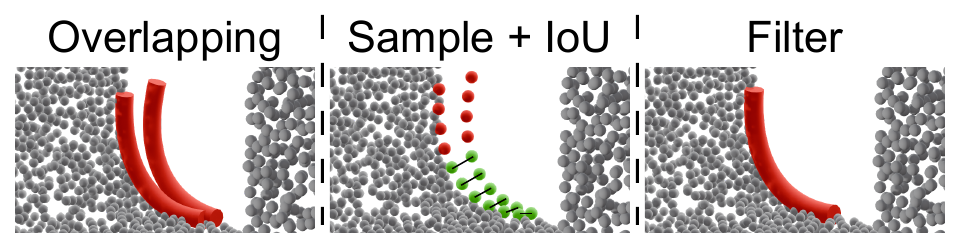}
        \caption{IoU Filter: removing overlapping or nearby curves by computing the IoU of overlapping points and discarding the curve with lower confidence.}
        \label{fig:pp_filter}
    \end{subfigure}
    
    \caption{Optional post-processing steps: (a) S\&F, (b) IoU Filter.}
    \label{fig:pp_pipeline}
\end{figure}

\paragraph{Loss function.} 
Let $i \in \{1, \dots, K\}$ be the index of a ground truth curve from the extended list of ground truth curves that includes the no-object class, and let $j = \sigma(i) \in \{1, \dots, K\}$ be the index of a prediction that was matched with $i$ in the matcher. The total loss for such a pair $(j,i)$ is defined as
\begin{align}
\label{eq:loss_total}
\begin{split}
    \mathcal{L}_\text{total}(j, i) =& \mathcal{L}_\text{CE}(\hat{p}_j, c_i) + \mathds{1}_{\{c_i\neq 0\}} \mathcal{L}_\text{param}(j, i) + \\
    & \mathds{1}_{\{c_i\neq 0\}} \mathcal{L}_\text{CD}(S^{c_i}_j, S^{c_i}_i),
\end{split}
\end{align}
where $\mathcal{L}_\text{CE}(\hat{p}_j, c_i)$ is the cross-entropy loss between the predicted class probabilities $\hat{p}_j$ and ground truth class $c_i$, $\mathcal{L}_\text{param}$ is the geometry-aware parametric curve loss, and $\mathcal{L}_\text{CD}$ is the Chamfer distance (CD) loss. Here, $S^{c_i}_j$ and $S^{c_i}_i$ denote the $64$ uniformly sampled 3D points from the curves of class $c_i$ using the respective parameters of $j$ and $i$.

Predictions matched to a placeholder ground truth instance with $c_i = 0$ contribute only to the class prediction branch through the cross entropy loss. 
We apply auxiliary losses after each decoder layer as in \cite{carion2020end, misra2021end} and since our dataset exhibits a class imbalance among curve types, we use class weights in the cross entropy loss. The weight computation and details of the CD loss are provided in \cref{apx:loss} and for completeness, CD is defined in \cref{apx:evaluation}.
\begin{figure}[b]
    \centering
    \includegraphics[width=1\linewidth]{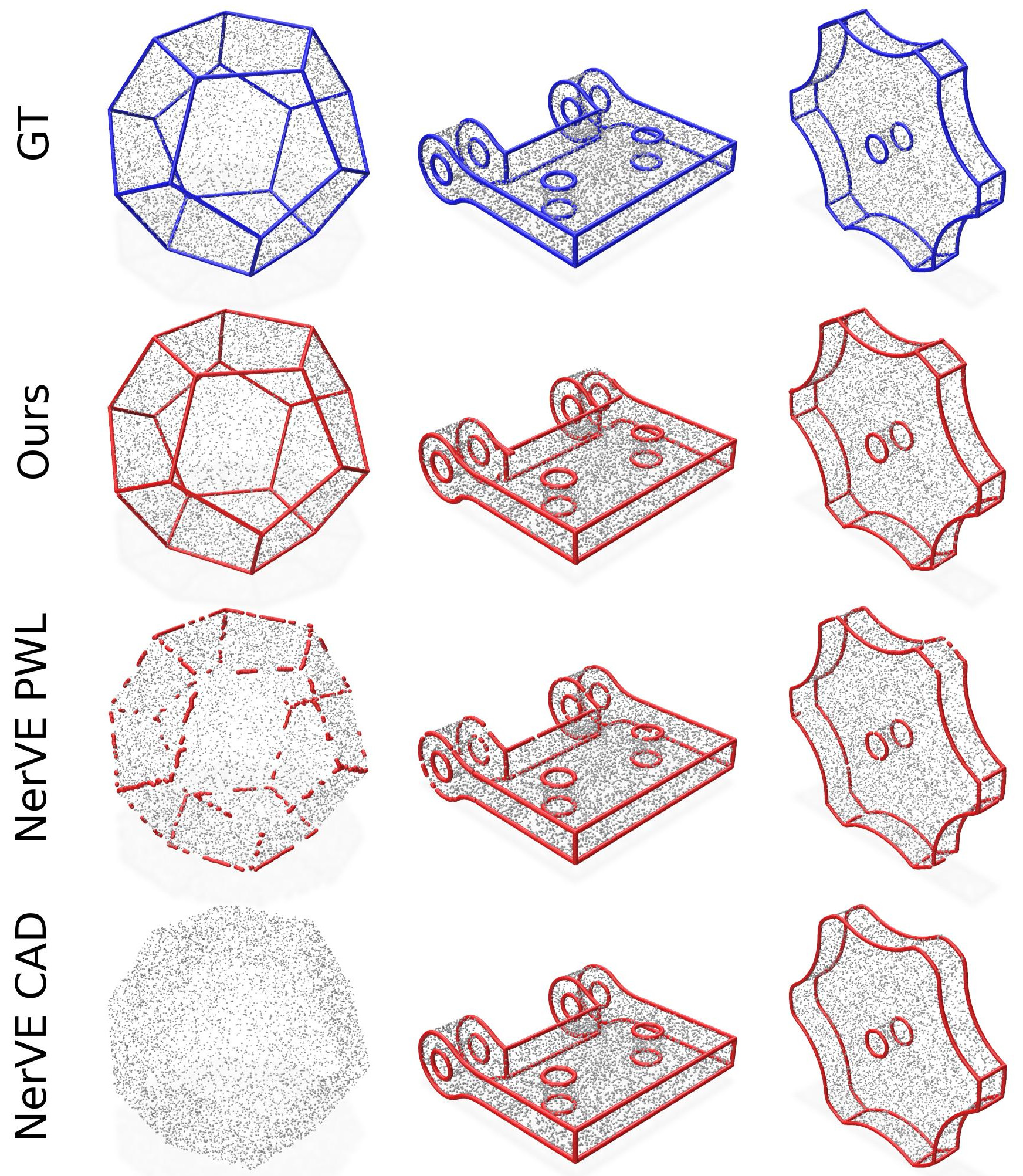}
    \caption{Qualitative comparison of predicted curves using $N = 32{,}768$. Our method and NerVE (CAD) predict parametric curves, while NerVE (PWL) produces piece-wise linear curves. NerVE (CAD) may produce invalid predictions, shown as empty.}

    \label{fig:pi3detr_comparison}
\end{figure}

\paragraph{Optional post-processing.}
We optionally apply two post-processing steps to refine PI3DETR predictions and suppress duplicates, with their impact analyzed in the ablation study (\cref{sec:ablation}). Snap \& Fit (S\&F) snaps predicted curves to the point cloud and refits them using its predicted class (\cref{fig:pp_snap}), while an IoU Filter removes overlapping curves of the same class based on sampled-point IoU and confidence (\cref{fig:pp_filter}). More details are provided in \cref{apx:post_processing}.

\subsection{Implementation Details}
PI3DETR is implemented in PyTorch \cite{paszke2019pytorch} and builds on the Transformer design of~\cite{carion2020end}. The SAModule \cite{qi2017pointnet++} downsamples each input cloud to $N'=2048$ points via FPS, computing $d=256$ features from $64$ neighbors within $\ell_2$-radius $0.2$. Features are projected to $d'=768$ and fed to $3$ Transformer encoder and $9$ decoder layers, each with $8$-head attention, a two-layer FFN (hidden dim $1024$), and dropout \cite{srivastava2014dropout} of $0.1$. Sine positional encodings \cite{vaswani2017attention} are applied to XYZ coordinates of encoder outputs and queries. Prediction heads are four-layer FFNs (dim $768$), except class, line length, and circle radius heads, which use three layers; all use LayerNorm \cite{ba2016layer} and ReLU \cite{agarap2018deep}.
Optimization uses AdamW \cite{loshchilov2017decoupled} with learning rate warm-up from $10^{-10}$ to $10^{-4}$ over $30$ epochs, then decayed to $10^{-5}$ at epoch $1230$, and gradient clipping at $\ell_2$-norm $0.2$. Models are trained for $1700$ epochs on $2$ NVIDIA TITAN RTX GPUs with an effective batch size of $128$ via gradient accumulation. Losses are normalized per-object. Data augmentation includes random point dropout ($85\%$ probability, retaining at least $20\%$ points) and per-instance random 3D rotation.

\section{Experiments}

\begin{figure*}
    \centering
    \includegraphics[width=1\linewidth]{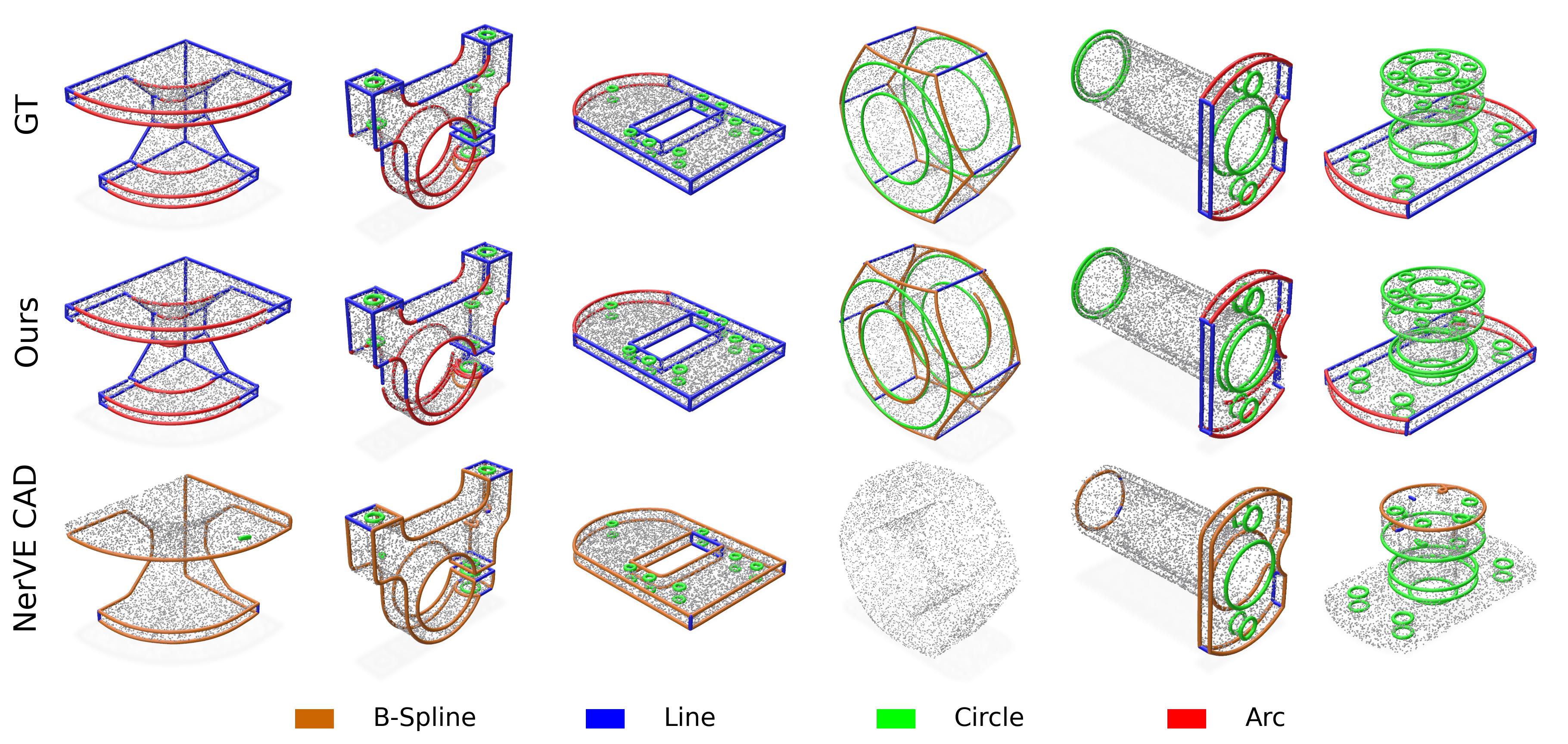}
    \caption{Qualitative comparison of curve instance predictions using $N = 32{,}768$. Our approach assigns the correct class at the appropriate locations, while NerVE often falls back to long B-splines and does not support arcs. Bézier (ours) and B-spline (NerVE) share the same color. NerVE may produce invalid predictions, shown as empty.}
    \label{fig:pi3detr_instance_comparison}
\end{figure*}

\label{sec:4_experiments}

We evaluate PI3DETR through both qualitative and quantitative experiments, including comparisons with 3D edge detection baselines, point sampling and noise robustness tests, and ablation studies. Beyond baseline comparisons, we demonstrate PI3DETR’s ability to directly regress parametric 3D curves, capturing rich geometric details both visually and quantitatively. 

\paragraph{Configuration.}
Our unseen test set consists of 924 samples (see \cref{sec:dataset} for details). The data is normalized to $[-1,1]^3$ for evaluation and uniformly formatted to ensure fairness and reproducibility. All runtime measurements are conducted on an NVIDIA RTX 4090 GPU. Unless otherwise specified, we use $N=32{,}768$ input points and $K=256$ point queries at inference (with training performed using $K=128$) together with Snap \& Fit (S\&F) post-processing, as this configuration provides the most stable performance across metrics according to our ablation study (\cref{sec:ablation}). 

\paragraph{Metrics.}
Since no prior method uses the same curve classes as PI3DETR, standard instance detection mean average precision (mAP) metrics are not directly applicable. Let $i \in \{1,\dots,M\}$ be ground truth curve and $j \in \{1,\dots,K\}$ be prediction curve indices. For each curve, we sample points $\tilde{S}^{c_i}_i$ (ground truth) or $\tilde{S}^{\hat{c}_j}_j$ (prediction) at intervals of $0.01$ along its type $c_i$ or predicted type $\hat{c}_j$. The aggregated point sets $\bigcup_{i=1}^M \tilde{S}^{c_i}_i$ and $\bigcup_{j=1}^K \tilde{S}^{\hat{c}_j}_j$ are used to compute Chamfer (CD) and Hausdorff (HD) distances, defined in \cref{apx:evaluation}. We also report mAP for completeness and future comparison, defining a match $(i,j)$ as a true positive if $c_i = \hat{c}_j$ and the CD between $\tilde{S}^{c_i}_i$ and $\tilde{S}^{\hat{c}_j}_j$ is below $0.005$.

\paragraph{Baseline settings.}
Since NerVE \cite{zhu2023nerve} established the state-of-the-art and outperformed prior methods, we primarily compare PI3DETR against NerVE. For completeness, we also attempted to set up PIE-NET \cite{wang2020pie} and DEF \cite{matveev2022def}, but PIE-NET has not released checkpoints for the second-stage curve fitting and DEF’s curve extraction scripts were error-prone and required non-trivial modifications to run reliably. We do not consider segmentation-based approaches that operate directly on the point cloud, as our focus is on parametric curve reconstruction methods. We evaluate NerVE using its released checkpoint and configuration. All reported metrics are computed in $[-1,1]^3$ space after transforming the outputs accordingly. NerVE predicts piece-wise linear (PWL) curves and fits parametric curves such as B-splines, line segments, and circles through a sensitive post-processing step that often fails to produce curves. In our tables, the number of such failures out of 924 test samples is shown as a superscript. For those cases, we exclude the failed samples from metric computation rather than penalizing them. This favors NerVE in comparison to PI3DETR, which always produced valid curves in the evaluations. Following our metrics protocol, we sample points at $0.01$ intervals along the parametric curves or PWLs to compute metrics. We denote the raw PWL outputs as NerVE PWL and the post-processed curves as NerVE CAD. Although PWL outputs are more comparable to segmentation and tend to achieve higher metric scores due to voxel anchoring, we still report them for completeness.

\begin{table}[b]
\centering

\begin{adjustbox}{max width=\linewidth}
\begin{tabular}{l|ccc}
\toprule
\textbf{Metric} & NerVE CAD & NerVE PWL & \textbf{Ours} \\
\midrule
CD $\downarrow$ & 0.0401 {\scriptsize ($\pm$ 0.20)} \textsuperscript{\textit{77}}  & 0.0046 {\scriptsize ($\pm$ 0.02)} & \textbf{0.0024} \textbf{\scriptsize ($\pm$ 0.02)} \\
HD $\downarrow$ & 0.2478 {\scriptsize ($\pm$ 0.35)} \textsuperscript{\textit{77}} & 0.1534 {\scriptsize ($\pm$ 0.17)} & \textbf{0.0635} \textbf{\scriptsize ($\pm$ 0.07)} \\
mAP $\uparrow$       & --    & --       & \textbf{0.8090} \\
\bottomrule
\end{tabular}
\end{adjustbox}

\caption{Quantitative comparison with $N = 32{,}768$. Invalid predictions shown in superscript.}
\label{tab:pi3detr_comparison}
\end{table}

\subsection{Comparison}
\label{sec:comparison}

We evaluate PI3DETR’s quantitative edge detection performance in \cref{tab:pi3detr_comparison}. With a CD of $0.0024$ and an HD of $0.0635$, our method clearly outperforms both NerVE CAD and NerVE PWL. The performance gap is especially pronounced for HD, where higher values indicate severe local mismatches. For NerVE CAD, $77$ cases fail during curve fitting from the PWL outputs. By omitting failed cases from NerVE CAD's evaluation, the comparison is biased in its favor, yet PI3DETR still significantly surpasses it. Beyond accuracy, we do not only outperform the PWL setting but also produce interpretable parametric curve instances directly, which the PWL outputs cannot provide. 

For qualitative evaluation, \cref{fig:pi3detr_comparison} illustrates PI3DETR’s strong prediction performance and also shows cases where NerVE’s post-processing has difficulties converting PWL outputs into valid CAD curves. In \cref{fig:pi3detr_instance_comparison}, we further present our instance-level predictions and directly usable CAD outputs. Unlike NerVE CAD, which often produces overly long and connected B-spline curves instead of lines or circles, PI3DETR infers semantically more meaningful and structured curve instances.

\begin{figure}[t]
        \centering
        \includegraphics[width=\linewidth]{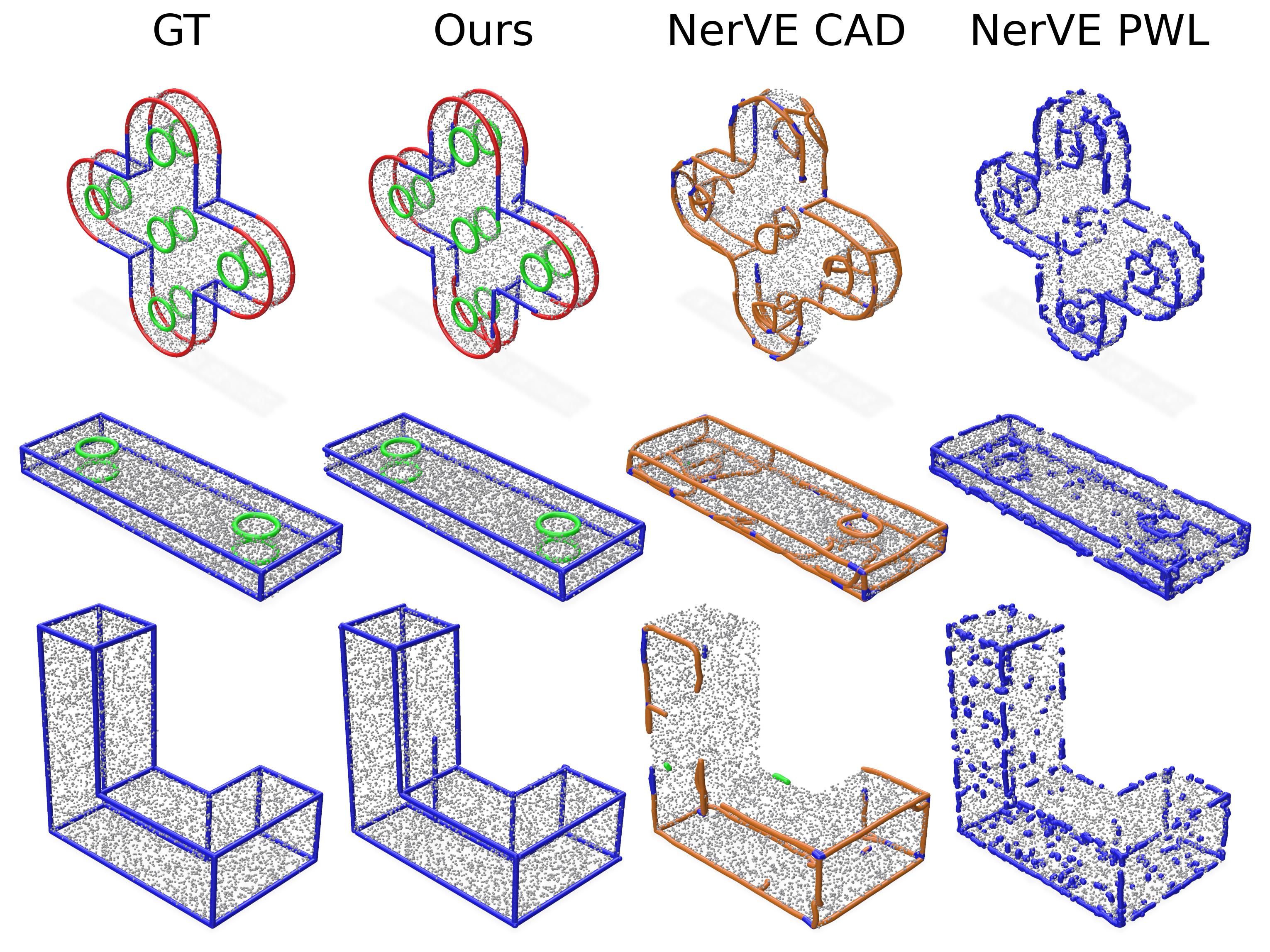}    
    \caption{Qualitative comparison on $N = 32{,}768$ affected by noise ($\eta = s/2e^2$) to evaluate robustness.}
    \label{fig:noise}
\end{figure}

\subsection{Stress Testing}

\paragraph{Point cloud density.}
To evaluate robustness, we vary point cloud density $N$ by randomly sub-sampling the test set (\cref{tab:stress_test}). PI3DETR consistently achieves the lowest CD and HD across all settings, with performance only degrading at $N = 4{,}096$. It surpasses all NerVE variants at $N = 8{,}192$ for CD and $N = 4{,}096$ for HD. In the sparsest setting, NerVE CAD produces 347 invalid outputs, while our method remains fully valid. To further validate our end-to-end design, we report results for the same sparse case without the S\&F post-processing, achieving a CD of $0.0049$ and an HD of $0.0841$, surpassing even the full pipeline.

\paragraph{Noise.}
To assess robustness against noise, we define three levels $\eta \in {\tfrac{s}{1e^3}, \tfrac{s}{5e^2}, \tfrac{s}{2e^2}}$, where $\eta$ is the standard deviation of Gaussian noise added to the 3D coordinates and $s$ denotes the largest axis-aligned span of the point cloud. Results in \cref{tab:stress_test_noise} show that PI3DETR consistently outperforms NerVE CAD across all noise levels for both CD and HD, and also surpasses NerVE PWL on HD. For CD, NerVE PWL performs better at lower noise levels, but PI3DETR achieves the best score once noise increases to $\eta=\tfrac{s}{2e^2}$. Without the post-processing step, PI3DETR attains a CD of $0.0138$ and an HD of $0.1986$, slightly weaker than the full pipeline yet still surpassing the NerVE baseline, further demonstrating its end-to-end robustness.

Although PI3DETR degrades relative to the noise-free setting, it remains consistently more robust than NerVE as noise increases. The stronger CD of NerVE PWL at low noise can be attributed to spurious PWLs detected in noisy point clouds, which lower CD but increase HD. This effect is visible in \cref{fig:noise}, which shows predictions of all methods for $\eta=\tfrac{s}{2e^2}$, and is consistent with the higher HD values reported in \cref{tab:stress_test_noise} or an invalid prediction percentage of 46\%. 

\begin{table}[t]
\centering

\begin{adjustbox}{max width=\linewidth}
\begin{tabular}{l|ccc}
\toprule
\textbf{N} & NerVE CAD & NerVE PWL & \textbf{Ours} \\
\midrule
& \multicolumn{3}{c}{\textit{Chamfer Distance (CD) $\downarrow$}} \\
\cmidrule(lr){2-4}
 32,768 & 0.0401 {\scriptsize ($\pm$ 0.20)} \textsuperscript{\textit{77}}& 0.0046 {\scriptsize ($\pm$ 0.02)} & \textbf{0.0024} \textbf{\scriptsize ($\pm$ 0.02)} \\
 16,384 & 0.1134 {\scriptsize ($\pm$ 0.43)} \textsuperscript{\textit{58}}& 0.0061 {\scriptsize ($\pm$ 0.02)} & \textbf{0.0025} \textbf{\scriptsize ($\pm$ 0.02)} \\
 8,192 & 0.2882 {\scriptsize ($\pm$ 0.60)} \textsuperscript{\textit{113}}& 0.0167 {\scriptsize ($\pm$ 0.04)} & \textbf{0.0027} \textbf{\scriptsize ($\pm$ 0.02)} \\
 4,096  & 0.4562 {\scriptsize ($\pm$ 0.68)} \textsuperscript{\textit{347}}& 0.0984 {\scriptsize ($\pm$ 0.27)} & \textbf{0.0050} \textbf{\scriptsize ($\pm$ 0.03)} \\
\midrule
& \multicolumn{3}{c}{\textit{Hausdorff Distance (HD) $\downarrow$}} \\
\cmidrule(lr){2-4}
 32,768 & 0.2477 {\scriptsize ($\pm$ 0.35)} \textsuperscript{\textit{77}}& 0.1534 {\scriptsize ($\pm$ 0.17)} & \textbf{0.0635} \textbf{\scriptsize ($\pm$ 0.07)} \\
 16,384 & 0.3961 {\scriptsize ($\pm$ 0.49)} \textsuperscript{\textit{58}}& 0.2008 {\scriptsize ($\pm$ 0.20)} & \textbf{0.0634} \textbf{\scriptsize ($\pm$ 0.07)} \\
 8,192  & 0.6436 {\scriptsize ($\pm$ 0.66)} \textsuperscript{\textit{113}}& 0.2987 {\scriptsize ($\pm$ 0.25)} & \textbf{0.0680} \textbf{\scriptsize ($\pm$ 0.08)} \\
 4,096  & 0.8665 {\scriptsize ($\pm$ 0.74)} \textsuperscript{\textit{347}}& 0.4918 {\scriptsize ($\pm$ 0.42)} &\textbf{0.0857} \textbf{\scriptsize ($\pm$ 0.10)} \\
 \bottomrule
\end{tabular}%
\end{adjustbox}
\caption{Robustness comparison under varying point cloud densities ($N$) from $32{,}768$ to $4{,}086$. Invalid predictions shown in superscript.}
\label{tab:stress_test}
\end{table}

\begin{table}[t]
\centering
\begin{adjustbox}{max width=\linewidth}
\begin{tabular}{l|ccc}
\toprule
\textbf{Noise} & NerVE CAD & NerVE PWL & \textbf{Ours} \\
\midrule
 & \multicolumn{3}{c}{\textit{Chamfer Distance (CD) $\downarrow$}} \\
\cmidrule(lr){2-4}
$\eta = s/1e^3$   & 0.0311 {\scriptsize ($\pm$ 0.21)} \textsuperscript{\textit{84}}& \textbf{0.0061 {\scriptsize ($\pm$ 0.02)}} & 0.0113 {\scriptsize ($\pm$ 0.06)} \\
$\eta = s/5e^2$   & 0.0194 {\scriptsize ($\pm$ 0.13)} \textsuperscript{\textit{199}} & \textbf{0.0121 {\scriptsize ($\pm$ 0.04)}} & 0.0129 {\scriptsize ($\pm$ 0.06)} \\
$\eta = s/2e^2$   & 0.0164 {\scriptsize ($\pm$ 0.04)} \textsuperscript{\textit{427}} & 0.0211 {\scriptsize ($\pm$ 0.05)} & \textbf{0.0134 {\scriptsize ($\pm$ 0.06)}} \\
\midrule
 & \multicolumn{3}{c}{\textit{Hausdorff Distance (HD) $\downarrow$}} \\
\cmidrule(lr){2-4}
$\eta = s/1e^3$   & 0.2306 {\scriptsize ($\pm$ 0.30)} \textsuperscript{\textit{84}} & 0.2530 {\scriptsize ($\pm$ 0.22)} & \textbf{0.1471 {\scriptsize ($\pm$ 0.11)}} \\
$\eta = s/5e^2$   & 0.2743 {\scriptsize ($\pm$ 0.25)} \textsuperscript{\textit{199}} & 0.3581 {\scriptsize ($\pm$ 0.23)} & \textbf{0.1684 {\scriptsize ($\pm$ 0.12)}} \\
$\eta = s/2e^2$   & 0.3086 {\scriptsize ($\pm$ 0.20)} \textsuperscript{\textit{427}} & 0.3874 {\scriptsize ($\pm$  0.23)} & \textbf{0.1946 {\scriptsize ($\pm$ 0.12)}} \\

\bottomrule
\end{tabular}%
\end{adjustbox}
\caption{Robustness comparison under varying point cloud noise ($\eta$) from $s/1e^3$ to $s/5e^2$ where $s$ is the point clouds span. Invalid predictions shown in superscript.}
\label{tab:stress_test_noise}
\end{table}

\subsection{Model Ablation}
\label{sec:ablation}
\paragraph{Point queries.}
Since PI3DETR employs non-parametric point queries \cite{misra2021end}, the number of queries $K$ can be varied at inference time. \cref{tab:ablation_k} presents an ablation study for $K \in \{128, 256, 512\}$, where results are reported after Snap \& Fit (S\&F) post-processing and runtimes are measured in seconds. For the main metrics, $K=128$ achieves the best overall mAP, $K=256$ yields the lowest Chamfer Distance (CD), and $K=512$ performs best in terms of Hausdorff Distance (HD). Notably, $K=128$ provides the strongest mAP across almost all primitives, with the exception of cubic Béziers where $K=256$ slightly outperforms it. Overall, $K=256$ emerges as the most stable configuration, balancing CD, HD, and mAP, whereas HD degrades significantly for $K=128$. In terms of efficiency, $K=128$ achieves the lowest runtime, while $K=512$ incurs the highest due to processing more output curves. The end-to-end runtime (inference plus S\&F) is $0.1551$s, $0.1923$s, and $0.2660$s for $K=128$, $256$, and $512$, respectively.

\paragraph{Post-processing.}
We also evaluate performance without post-processing. As shown in \cref{tab:ablation_post}, even without refinement PI3DETR achieves strong results. Applying S\&F offers slight improvements in CD and mAP, with a more noticeable gain in HD, suggesting that S\&F effectively mitigates large errors. Interestingly, the IoU-based filtering step does not improve results, which may be explained by the frequent occurrence of thin structures in the dataset, where closely aligned curves risk being mistakenly suppressed.

\begin{table}[t]
\centering
\begin{adjustbox}{max width=\linewidth}
\begin{tabular}{l|ccc}
\hline
 & K=128 & K=256 & K=512 \\
\hline
 & \multicolumn{3}{c}{\textit{Geometric Quality (CD, HD) $\downarrow$}} \\
\cmidrule(lr){2-4}
CD $\downarrow$  & 0.0026 {\scriptsize ($\pm$ 0.02)} & \textbf{0.0024 {\scriptsize ($\pm$ 0.02)}} & 0.0025 {\scriptsize ($\pm$ 0.02)} \\
HD $\downarrow$  & 0.0671 {\scriptsize ($\pm$ 0.08)} & 0.0635 {\scriptsize ($\pm$ 0.08)} & \textbf{0.0629 {\scriptsize ($\pm$ 0.07)}} \\
\hline
 & \multicolumn{3}{c}{\textit{Detection (AP) $\uparrow$}} \\
\cmidrule(lr){2-4}
Bézier $\uparrow$    & 0.6156 & \textbf{0.6184} & 0.5960 \\
Line  $\uparrow$     & \textbf{0.9416} & 0.9192 & 0.9203 \\
Circle $\uparrow$    & \textbf{0.9045} & 0.8871 & 0.8824 \\
Arc $\uparrow$ & \textbf{0.8373} & 0.8114 & 0.8063 \\
\cmidrule(lr){2-4}
mAP $\uparrow$  & \textbf{0.8247} & 0.8090 & 0.8012 \\
\hline
 & \multicolumn{3}{c}{\textit{Timing (sec) $\downarrow$}} \\
\cmidrule(lr){2-4}
Model $\downarrow$ & \textbf{0.1128 {\scriptsize ($\pm$ 0.02)}} & 0.1181 {\scriptsize ($\pm$ 0.03)} & 0.1337 {\scriptsize ($\pm$ 0.03)} \\
S\&F $\downarrow$  & \textbf{0.0423 {\scriptsize ($\pm$ 0.05)}} & 0.0742 {\scriptsize ($\pm$ 0.10)} & 0.1323 {\scriptsize ($\pm$ 0.19)} \\
\hline
\end{tabular}
\end{adjustbox}
\caption{Evaluation metrics for different numbers of point queries $K$ and $N = 32{,}768$. All results are reported with Snap \& Fit (S\&F) post-processing. Runtimes are measured in seconds.}
\label{tab:ablation_k}
\end{table}

\begin{table}[t]
\centering
\begin{adjustbox}{max width=\linewidth}
\begin{tabular}{@{}l|ccc@{}}
\toprule
\textbf{Configuration} & CD $\downarrow$ & HD $\downarrow$ &  mAP  $\uparrow$ \\
\midrule
No Post-processing                 & 0.0025 \scriptsize ($\pm$ 0.02) & 0.0663 \scriptsize ($\pm$ 0.07) & 0.8043 \\
+ Snap \& Fit             & \textbf{0.0024 \scriptsize ($\pm$ 0.02)} & \textbf{0.0635 \scriptsize ($\pm$ 0.07)}  & \textbf{0.8090} \\
+ IoU Filter       & \textbf{0.0024 \scriptsize ($\pm$ 0.02)} & 0.0636 \scriptsize ($\pm$ 0.07)  & 0.7981 \\
\bottomrule
\end{tabular}
\end{adjustbox}
\caption{Ablation study of the optional post-processing methods evaluated with $K=256$ queries and $N = 32{,}768$.}
\label{tab:ablation_post}
\end{table}

\subsection{Real World Evaluation}
\begin{figure}[t]
    \centering
    \includegraphics[width=1\linewidth]{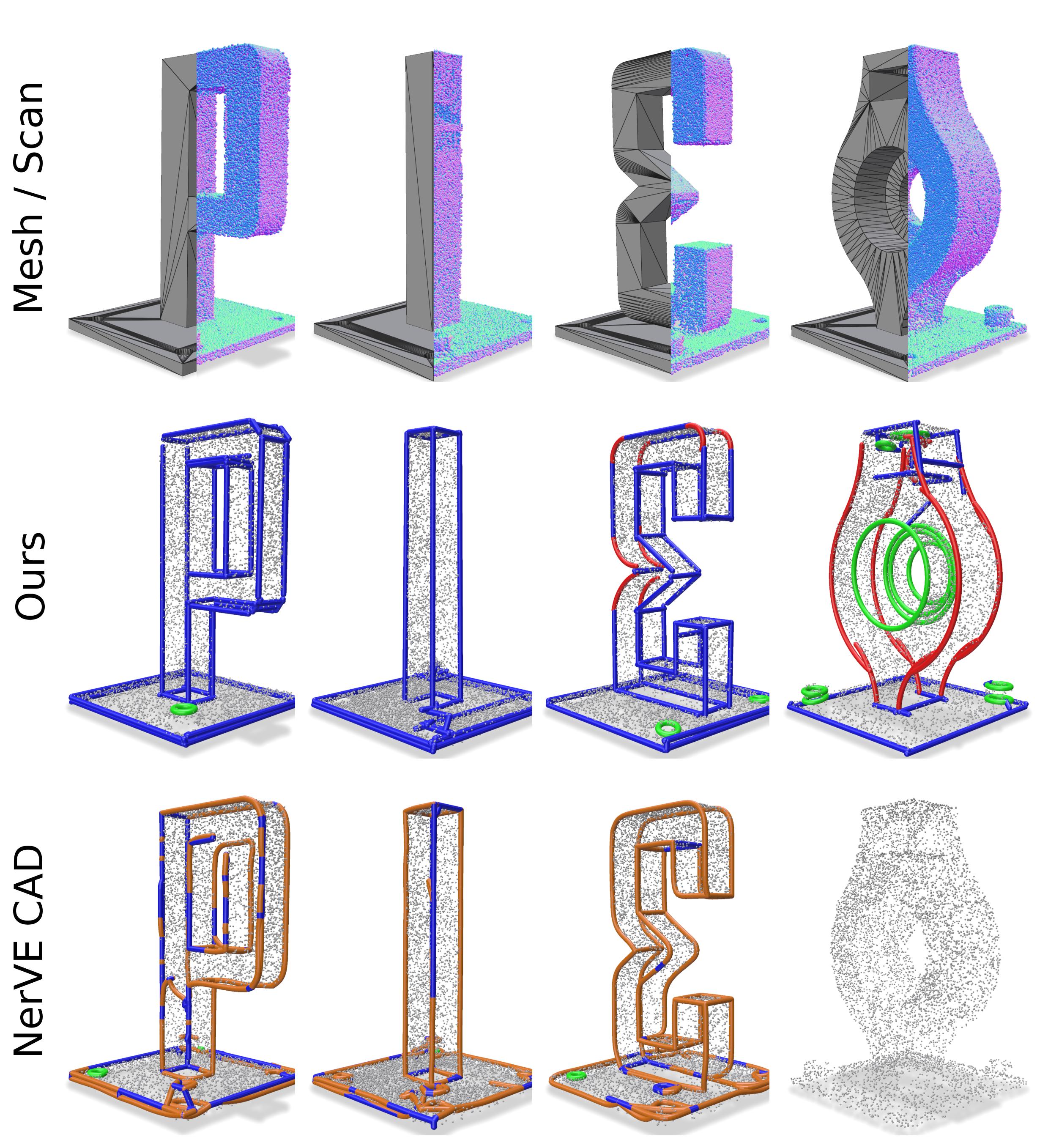}
    \caption{Qualitative comparison of edge detection results on real-world scanned CAD models unseen during training.}
    \label{fig:real_point_clouds}
\end{figure}

The ABC dataset does not contain any real-world object scans in the form of point clouds. To demonstrate our model’s generalization to unseen, real-world data, we generated several CAD models of varying complexity that were not part of the ABC dataset. These models were 3D-printed and scanned with a 7-DOF robotic system equipped with a structured light sensor. Data acquisition followed the inspection plan proposed by Staderini \etal \cite{staderini2023surface,staderini2024visual}. The resulting dense point clouds were randomly sub-sampled to $N = 32{,}768$ points for inference. \cref{fig:real_point_clouds} shows the original mesh, the scanned point cloud, and our results, producing curves consistent with how the edges would be represented in a CAD model. NerVE fails on the right mesh and produces inconsistent, irregular curves on the others.

\section{Conclusion}
\label{sec:5_introduction}
We presented PI3DETR, an end-to-end framework for detecting differently parameterized 3D curve instances in point clouds in a single forward pass. By incorporating a geometry-aware matching strategy, PI3DETR removes the need for intermediate representations or multi-stage processing and can optionally use simple post-processing steps. We achieve state-of-the-art performance, lowering CD on the ABC dataset by almost half compared to NerVE, and is particularly robust on sparse point clouds. 
PI3DETR’s limitations are the need for extensive training to achieve optimal results, the lack of a fallback for misclassified curves, and the challenge of directly regressing curves without explicit point associations. Nevertheless, its simplicity makes it a powerful foundation for future advances in 3D curve detection and geometric reasoning, paving the way toward more structured and intelligent 3D understanding.

{
    \small
    \bibliographystyle{ieeenat_fullname}
    \bibliography{main}
}

\clearpage
\appendix
\setcounter{page}{1}
\maketitlesupplementary

\section{Dataset}
\label{apx:dataset}

\paragraph{Dataset filtering.}To ensure consistent supervision and stable evaluation, we apply a filtering protocol to the ABC dataset \cite{koch2019abc}. The dataset is organized in chunks of 10,000 models, and we use the first five chunks. Following \cite{wang2020pie}, which reports that over $95\%$ of curves are lines, circles, or B-splines and discards models with more than $300.000$ vertices, we restrict our study accordingly. We add arcs as a separate class to capture non-closed circles and B-splines that are well represented by circular arcs, and we model remaining B-splines as cubic Bézier curves for better parameterization. We retain only models containing circles, cubic Bézier curves, line segments, or arcs. Models with more than $100$ curves are removed. We also discard models containing circles with radius below $1\%$ and lines shorter than $0.01\%$ of the bounding box diagonal. Finally, we manually remove erroneous models, bevel heavy shapes, and models containing only circles. The resulting 10,240 samples are split into $80\%$for training, $10\%$ for validation, and $10\%$ for testing.

\section{PI3DETR Architecture}
This section details PI3DETR’s architecture, providing a complete explanation of the geometry-aware matching strategy, the computation of loss weights, the definitions of the loss functions, and the evaluation metrics.
\label{apx:architecture}

\subsection{Geometry-Aware Matching}
\label{apx:matching}
This section provides a slightly more detailed explanation of the matching procedure. To train the model, we need to match the set of predicted 3D curves to the ground truth curves. Since PI3DETR computes parameters for all four curve types for each decoder output embedding in $\{o^\text{f}_j\}_{j=1}^K$, we eventually obtain $4K$ curves that are handled by the geometry-aware matcher. Given a prediction with index $j \in \{1,\dots,K\}$ and a ground truth curve with index $i \in \{1,\dots,M\}$, the matching cost $\mathcal{C}_\text{match}$ for a pair $(j,i)$ of curves is
\begin{align}
    \mathcal{C}_\text{match}(j, i) = -\hat{p}_j(c_i) + \mathcal{L}_\text{param}(j, i),
\end{align}
where the class cost $-\hat{p}_j(c_i)$ is the negative predicted probability of query $j$ being class $c_i$ \cite{carion2020end}, and $\mathcal{L}_\text{param}(j, i)$ is the geometric cost (\cref{apx:loss_param}) selected accordingly to \cref{eq:loss_param}. 
We use cost and loss interchangeably for function reuse. Unlike the final loss in \cref{eq:loss_total}, however, the matcher does not compute gradients.
\begin{align}
\label{apx:loss_param}
\begin{split}
    \mathcal{L}&_\text{param}(j, i) = \\
    &\mathds{1}_{\{c_i=1\}} 
    \mathcal{L}_{\text{seq}}(
    \hat{\textbf{B}}_j, 
    \textbf{B}_i) +
    \mathds{1}_{\{c_i=2\}} 
    \mathcal{L}_\text{hybrid}
    (\hat{\mathbf{L}}_j
    \textbf{L}_i, 
    ) + \\
    & \mathds{1}_{\{c_i=3\}} 
    \mathcal{L}_\text{hybrid}
    (\hat{\mathbf{C}}_j, 
    \mathbf{C}_i,
    ) +
    \mathds{1}_{\{c_i=4\}} 
    \mathcal{L}_{\text{seq}}
    (\hat{\textbf{A}}_j, \textbf{A}_i)
\end{split}
\end{align}

The loss function $\mathcal{L}_{\text{seq}}$ in \cref{eq:loss_sequence} measures the discrepancy between two curves parameterized as ordered point sequences (hence $\mathcal{L}_\text{seq}$, encompassing both cubic Bézier curves and arcs. Given a predicted sequence $\hat{\mathbf{B}}$ and ground truth sequence $\mathbf{B}$, the loss is defined as
\begin{align}
\label{apx:loss_sequence}
\mathcal{L}_{\text{seq}}(\hat{\mathbf{B}}, \mathbf{B}) = \min \{ \| \hat{\mathbf{B}} - \mathbf{B} \|_1, \, \| \mathrm{rev}(\hat{\mathbf{B}}) - \mathbf{B} \|_1 \},
\end{align}
and represents an order-invariant \cite{polyroad} $\ell_1$-loss. In \cref{eq:loss_sequence}, $\mathrm{rev}(\cdot)$ denotes sequence reversal. Specifically, for a cubic Bézier curve 
$\hat{\mathbf{B}} = [\hat{\mathbf{b}}^\mathrm{s}, \hat{\mathbf{b}}^\mathrm{c1}, \hat{\mathbf{b}}^\mathrm{c2}, \hat{\mathbf{b}}^\mathrm{e}]$, we have 
$\mathrm{rev}(\hat{\mathbf{B}}) = [\hat{\mathbf{b}}^\mathrm{e}, \hat{\mathbf{b}}^\mathrm{c2}, \hat{\mathbf{b}}^\mathrm{c1}, \hat{\mathbf{b}}^\mathrm{s}]$. For an arc 
$\hat{\mathbf{A}} = [\hat{\mathbf{a}}^\mathrm{s}, \hat{\mathbf{a}}^\mathrm{m}, \hat{\mathbf{a}}^\mathrm{e}]$, the reversed sequence is 
$\mathrm{rev}(\hat{\mathbf{A}}) = [\hat{\mathbf{a}}^\mathrm{e}, \hat{\mathbf{a}}^\mathrm{m}, \hat{\mathbf{a}}^\mathrm{s}]$.
This formulation enforces invariance to the curve parameterization direction, reflecting the geometric equivalence of both orders.

We define a hybrid parameter loss $\mathcal{L}_\text{hybrid}$ in \cref{eq:loss_line_circle} to jointly handle line segments and circles. Both share parameters consisting of a 3D point, a direction or normal vector, and a scalar (length or radius). The loss combines an $\ell_1$-loss on the scalar with an $\ell_1$-loss on the point and vector pair. Since the direction vector is ambiguous up to sign, meaning $\mathbf{l}^\text{v}$ and $-\mathbf{l}^\text{v}$ represent the same orientation, the loss evaluates both options and uses the minimum to ensure invariance to vector direction, as shown in \cref{fig:curve_representations}.
Formally, the loss is defined as
\begin{align}
\label{apx:loss_line_circle}
\begin{split}
    \mathcal{L}_\text{hybrid}(\hat{\mathbf{L}}, \mathbf{L}) = \min \{ 
    \| (\hat{\mathbf{l}}^\text{m}, \hat{\mathbf{l}}^\text{v}) - (\mathbf{l}^\text{m}, \mathbf{l}^\text{v}) \|_1, \, \\
    \| (\hat{\mathbf{l}}^\text{m}, \hat{\mathbf{l}}^\text{v}) - (\mathbf{l}^\text{m}, -\mathbf{l}^\text{v}) \|_1 
    \} + \| \hat{l}^\text{r} - l^\text{r} \|_1 \,.
\end{split}
\end{align}
This loss formulation applies analogously to a circle $\mathbf{C} = [\mathbf{c}^m, \mathbf{c}^v, c^r]$ parameterized by center point, unit normal vector, and radius. The sign ambiguity of the normal vector is handled identically, making $\mathcal{L}_\text{hybrid}$ a unified loss function for both lines and circles.

Based on the defined cost function, we compute a cost-minimal bipartite matching between the $K$ predictions and $M \leq K$ ground truth curves using the Hungarian algorithm \cite{kuhn1955hungarian}, following \cite{carion2020end, matveev2022def}. The matching is represented by a permutation $\sigma \in \mathfrak{S}_K$ that assigns each ground truth index $i \in \{1, \dots, M\}$ to a unique prediction $j = \sigma(i)$. For the $K - M$ unmatched predictions, no-object class placeholders with $c_{\sigma(i)} = 0$ are introduced and matched accordingly, extending $i$ to the full set $\{1, \dots, K\}$ including these padded entries.

\begin{figure*}
    \centering
    \includegraphics[width=0.9\linewidth]{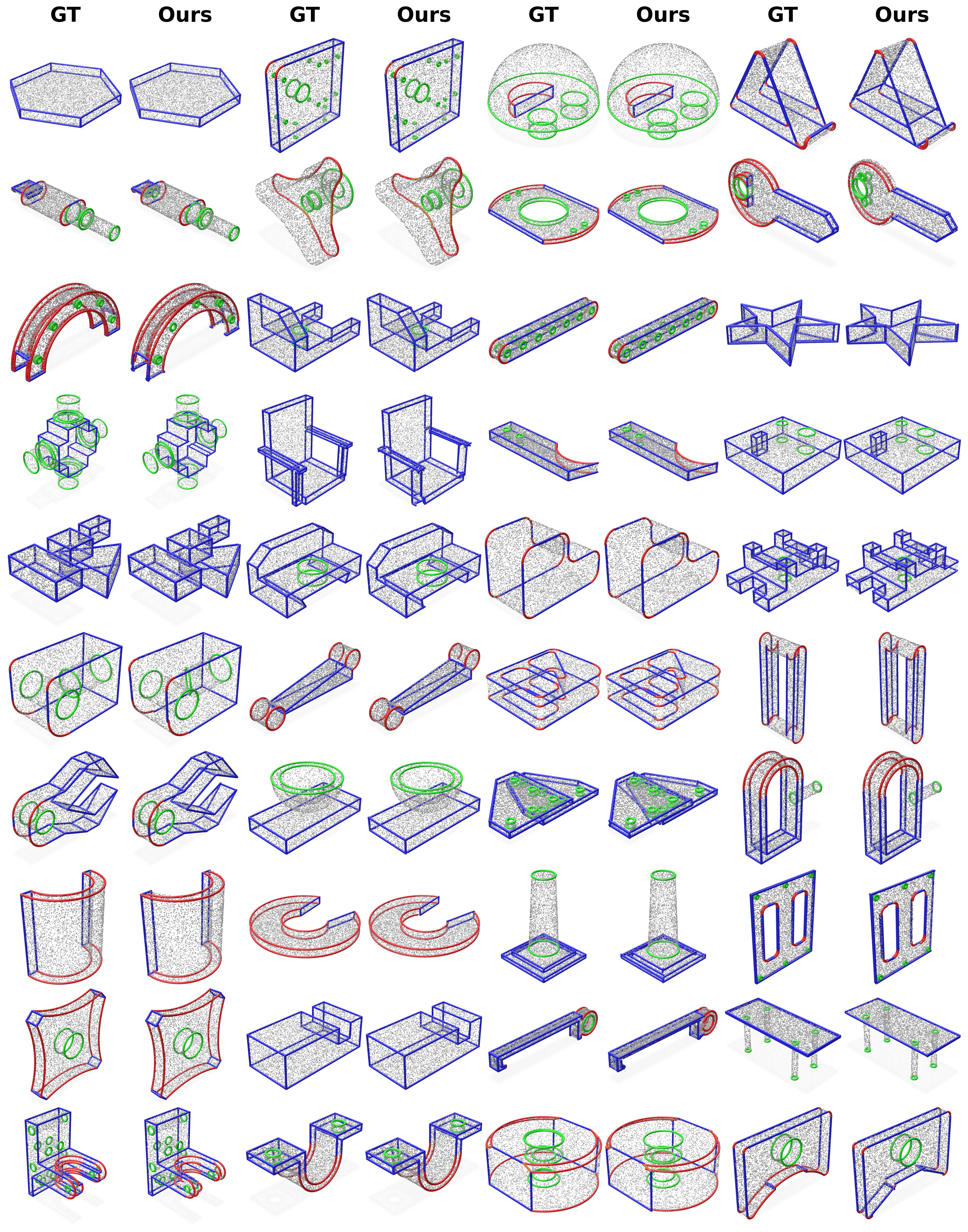}
    \caption{Qualitative visualization of PI3DETR’s performance on various models from the ABC Dataset, shown in comparison to the corresponding ground truth. Each example is evaluated with a point sampling of $N = 32{,}768$ using best-performing trained model.}
    \label{fig:big_comparison}
\end{figure*}

\subsection{Optional Post-Processing}
\label{apx:post_processing}
Since PI3DETR directly regresses curves in 3D space, predictions may occasionally be slightly offset or duplicated. To mitigate this, we introduce two post-processing steps that are optional but can further enhance performance when properly parameterized as shown in the ablation study in \cref{sec:ablation}. The first, Snap \& Fit (S\&F), aligns predicted curves with the underlying point cloud and enforces conformity to their geometric class (\cref{fig:pp_snap}). For each predicted curve $j$, points are sampled along the curve and matched to nearest neighbors in the point cloud, after which the curve is refit using a class-specific fitting method determined by the predicted class $\hat{c}_j = \arg\max_{k \in [0,4]} \hat{p}_j^{(k)}$. The second step, IoU Filter, removes duplicate or overlapping curves of the same class (\cref{fig:pp_filter}). Points are sampled along each curve, and pairwise distances between the sampled points of two curves are computed. Points within a predefined distance tolerance are considered overlapping, and the Intersection over Union (IoU) of these overlapping sets is calculated. If the IoU exceeds a threshold, the curve with lower confidence is discarded. In our experiments we set the IoU threshold to $0.6$ and the distance tolerance to $0.01$.

\subsection{Loss}
\label{apx:loss}
\paragraph{Class weight computation.}  
The class weights for the cross-entropy loss are computed from the distribution of curve classes in the training dataset. 
We use the class encoding \(c = 0\) for the no-object class, \(c = 1\) for cubic B\'ezier, \(c = 2\) for line segment, \(c = 3\) for circle, and \(c = 4\) for arc. 
The sample counts for the non-empty classes are \(n_1 = 11347\), \(n_2 = 200751\), \(n_3 = 34672\), and \(n_4 = 37528\). 
The count for the no-object class is computed as \(n_0 = K_\mathrm{total} - (n_1 + n_2 + n_3 + n_4)\), where \(K_\mathrm{total} = K \times N_\mathrm{train}\) is the total number of predictions in the dataset, \(K = 128\) is the number of decoder queries per sample, and \(N_\mathrm{train} = 8389\) is the number of training samples. 
Unnormalized weights are given by \(w_c' = 1 / \sqrt{n_c}\) and are then normalized to sum to one via \(w_c = w_c' / \sum_{c'=0}^4 w_{c'}'\). 
This results in the final weight vector 
\([0.048,\ 0.403,\ 0.096,\ 0.231,\ 0.222]\),
which is applied in the class-weighted cross-entropy loss.

\paragraph{Chamfer loss computation.}  
Both $S^{c_i}_i$ (ground truth) and $S^{c_i}_j$ (prediction) contain 64 uniformly sampled points along a curve of class $c_i$, where sampling is performed on the respective curve type using the parameters of $i$ for $S^{c_i}_i$ and the parameters of $j$ for $S^{c_i}_j$.
For example, if $c_i = 4$ (Bézier), we use the Bézier control points $\mathbf{B}_i$ and $\hat{\mathbf{B}}_j$ to sample the corresponding point sets.
If $c_i = 2$ (line segment), we take the line parameters $\mathbf{L}_i$ and $\hat{\mathbf{L}}_j$ to form $S^{1}_i$ and $S^{1}_j$. 
If $c_i = 3$ (circle), the sets $S^{3}_i$ and $S^{3}_j$ are generated from the circle parameters $\mathbf{C}_i$ and $\hat{\mathbf{C}}_j$. 
Finally, if $c_i = 2$ (arc), we use the arc parameters $\mathbf{A}_i$ and $\hat{\mathbf{A}}_j$. 
In each case, sampling is performed uniformly along the specified curve type, ensuring that both sets contain exactly 64 points.

\section{Evaluation}
\label{apx:evaluation}
We define the Chamfer distance between two 3D point sets $X$ and $Y$ as  
\begin{align}
\label{eq:chamfer}
\begin{split}
    \mathrm{CD}(X, Y) =& \frac{1}{|X|} \sum_{\mathbf{x} \in X} \min_{\mathbf{y} \in Y} \|\mathbf{x} - \mathbf{y}\|_2^2+\\
    &\frac{1}{|Y|} \sum_{\mathbf{y} \in Y} \min_{\mathbf{x} \in X} \|\mathbf{x} - \mathbf{y}\|_2^2.
\end{split}
\end{align}

We define the Hausdorff distance between two 3D point sets X and Y as
\begin{align}
\label{eq:hausdorff}
\begin{split}
    \mathrm{HD}(X, Y) = \frac{1}{2} \big(\max_{\mathbf{x} \in X} \min_{\mathbf{y} \in Y} \|\mathbf{x} - \mathbf{y}\|_2+\max_{\mathbf{y} \in Y} \min_{\mathbf{x} \in X} \|\mathbf{x} - \mathbf{y}\|_2 \big).
\end{split}
\end{align}

\subsection{Runtime}
\label{apx:runtime}

\begin{table}[H]
\centering
\begin{tabular}{l|cc}
\toprule
$\mathbf{N}$ & Model $\downarrow$ & Post-processing $\downarrow$ \\
\hline
32,768  & 0.1181 {\scriptsize ($\pm$ 0.02)} & 0.0705 {\scriptsize ($\pm$ 0.10)} \\
16,384  & 0.0723 {\scriptsize ($\pm$ 0.02)} & 0.0693 {\scriptsize ($\pm$ 0.10)} \\
8,192  & 0.0502 {\scriptsize ($\pm$ 0.02)} & 0.0696 {\scriptsize ($\pm$ 0.10)} \\
4,096  & 0.0359 {\scriptsize ($\pm$ 0.02)} & 0.0735 {\scriptsize ($\pm$ 0.10)} \\

\bottomrule
\end{tabular}
\caption{Inference and post-processing time measurement in seconds (NVIDIA RTX 4090) using different input point counts $N$ and $K = 256$.}
\label{tab:performance}
\end{table}

\end{document}